\documentclass[a4,a4wide]{article}
\usepackage{aaai}
\usepackage{times}
\usepackage{helvet}
\usepackage{courier}
\newtheorem{definition}{Definition}
\newtheorem{proposition}{Proposition}
\newenvironment{example}{{\vspace*{12pt}\noindent \bf Example} \ }{\vspace*{12pt}}

\begin{document}
\nocopyright

\title{Belief Revision and Trust}
\author{Aaron Hunter\\
British Columbia Institute of Technology\\
Burnaby, Canada\\
\texttt{aaron\_hunter@bcit.ca}
}
\maketitle
\begin{abstract}
Belief revision is the process in which an agent incorporates a new piece of information together with a pre-existing set of beliefs.   When the new information comes in the form of a report from another agent, then it is clear that we must first determine whether or not that agent should be trusted.  In this paper, we provide a formal approach to modeling trust as a pre-processing step before belief revision.  We emphasize that trust is not simply a relation between agents; the trust that one agent has in another is often restricted 
to a particular domain of expertise.  We demonstrate that this form of trust can be captured by associating a state-partition with each agent, then relativizing all reports to this state partition before performing belief revision.  In this manner, we incorporate only the part of a report that falls under the perceived domain of expertise of the reporting agent.  Unfortunately, state partitions based on expertise do not allow us to compare the relative strength of trust held with respect to different agents.  To address this problem, we introduce pseudometrics over states to represent differing degrees of trust.  This allows us to incorporate simultaneous reports from multiple agents in a way that ensures the most trusted reports will be believed.
\end{abstract}

\section{Introduction}
\label{sec:introduction}
The notion of trust must be addressed in many agent communication systems.  In this paper, we consider one isoloated aspect of trust: the manner in which trust impacts the process of belief revision.  Some of the most influential approaches to belief revision have used the simplifying assumption that all new information must be incorporated; however, this is clearly untrue in cases where information comes from an untrusted source. 
In this paper, we are concerned with the manner in which an agent uses an external notion of trust in order to determine how new information should be integrated with some pre-existing set of beliefs.    

Our basic approach is the following.  We introduce a simple model of trust that allows an agent to determine if a source can be trusted to distinguish between different pairs of states.  We use this notion of trust as a precursor to belief revision.  Hence, before revising by a new formula, an agent first determines to what extent the source of the information can be trusted.  In many cases, the agent will only incorporate ``part'' of the formula into their beliefs.  We then extend our model of trust to a more general setting, by introducing quantitative measures of trust that allow us to compare the degree to which different agents are trusted.  Fundamental properties are introduced and established, and applications are considered.

\section{Preliminaries}
\subsection{Intuition}
It is important to note that an agent typically does not trust another agent universally.  As such, we will not apply the label ``trusted'' to another agent; instead, we will say that an agent is trusted with respect to a certain domain of knowledge.  This is further complicated by the fact that there are different reasons that an agent may not be trusted.  For example, an agent might not be trusted due to their perceived knowledge of a domain.  In other cases, an agent might not be trusted due to their perceived dishonesty, or bias.  In this paper, our primary focus is on trust as a function of the perceived expertise of other agents.  Towards the end, we briefly address the different formal mechanisms that would be required to deal with deceit.

\subsection{Motivating Example}
We introduce a motivating example in commonsense reasoning where  an agent must rely on an informal notion of trust in order to inform rational belief change;  we will return to this example periodically as we introduce our formal model.   

Consider an agent that visits a doctor, having difficulty breathing.  Incidentally, the agent is wearing a necklace that prominently features a jewel on a pendant.  During the examination, the doctor checks the patient's throat for swelling or obstruction; at the same time, the doctor happens to look at the necklace.  Following the examination, the doctor tells the patient ``you have a viral infection in your throat - and by the way, you should know that the jewel in your necklace is not a diamond.''

The important part about this example is the fact that the doctor provides information about two distinct domains:  human health and jewelry.  In practice, a patient is very likely to trust the doctor's diagnosis about the viral infection.  On the other hand, the patient really has very little reason to trust the doctor's evaluation of the necklace.  We suggest that a rational agent should actually incorporate the doctor's statement about the infection into their own beliefs, while essentially ignoring the comment on the necklace.  This approach is dictated by the kind of trust that the patient has in the doctor.  Our aim in this paper is to formalize this kind of ``localized'' domain-specific trust, and then demonstrate how this form of trust is used in practice to inform belief revision.

\subsection{Trust}
Trust consists of two related components.  First, we can think of trust in terms of how likely an agent is to believe what another agent says.  Alternatively, we can think of trust in terms of the degree to which an agent is likely to allow another to perform actions on their behalf. In this paper, we will be concerned only with the former.

A great deal of existing work on trust focuses on the manner in which an agent develops a {\em reputation} based on past behaviour.  A brief survey of reputation systems is given in \cite{Huynh06}.   Reputation systems can be used to inform the allocation of tasks \cite{Ramchurn09}, or to avoid deception \cite{Salehi09}.  The model of trust presented in this paper is not intended to be an alternative to existing reputation systems;  we are not concerned with the manner in which an agent learns to trust another.  Instead, our focus is simply on developing a suitable model of trust that is expressive enough to inform the process of {\em belief revision}.
The manner in which this model of trust is developed over time is beyond the scope of this paper.

\subsection{Belief Revision}
{\em Belief revision} refers to the process in which an agent must integrate new information with some pre-existing beliefs about the state of the world.  One of the most influential approaches to belief revision is the AGM approach, in which an agent incorporates the new information while keeping as much of the intial belief state as consistently possible \cite{AlchourronGardenforsMakinson85}.  

This approach was originally defined with respect to a finite set $P$ of propositional variables representing properties of the world.  A {\em state} is a propositional interpretation over $P$, representing a possible state of the world.   A {\em belief set} is a deductively closed set of formulas, representing the beliefs of an agent.  Since $P$ is finite, it follows that every belief set defines a corresponding {\em belief state}, which is the set of states that an agent considers to be possible.  A revision operator is a function that takes a belief set and a formula as input, and returns a new belief set.  An AGM revision operator is a revision operator that satisfies the AGM postulates, as specified in \cite{AlchourronGardenforsMakinson85}.  

It turns out that every AGM revision operator is characterized by a total pre-order over possible worlds.  To be more precise, a {\em faithful assignment} is a function that maps each belief set to a total pre-order over states in which the models of the belief set are the minimal states.  When an agent is presented with a new formula $\phi$ for revision, the revised belief state is the set of all minimal models of $\phi$ in the total pre-order given by the faithful assignment.  We refer the reader to \cite{KatsunoMendelzon92} for a proof of this result, as well as a complete description of the implications.  For our purposes, we simply need to know that each AGM revision operator necessarily defines a faithful assignment.

\section{A Model of Trust}
\subsection{Domain-Specific Trust}
Assume we have a fixed propositional signature $\mathbf{F}$ as well as a set of agents $\mathbf{A}$.  
For each $A\in\mathbf{A}$, let $Bel_A$ denote  a deductively closed set of formulas over $\mathbf{F}$ called the {\em belief set} of $A$.  For each $A$, let $*_A$  denote an AGM revision operator that intuitively captures the way that the agent $A$ revises their beliefs when presented with new information.  This revision operator represents sort of an ``ideal'' revision situation, in which $A$ has complete trust in the new information.  We want to modify the way this operator is used, by adding a representation of the extent to which $A$ trusts each other agent $B\in\mathbf{A}$ over ${\mathbf F}$.  

We assume that all new information is reported by an agent, so each formula for revision can be labelled with the name of the reporting agent.\footnote{This is not a significant restriction.  In domains involving sensing or other forms of discovery, we could simply allow an agent $A$ to self-report information with complete trust.}  At this point, we are not concerned with degrees of trust or with resolving conflicts between different sources of information.  Instead, we start with a binary notion of trust, where $A$ either trusts $B$ or does not trust $B$ with respect to a particular domain of expertise.

We encode trust by allowing each agent $A$ to associate a partition $\Pi^B_A$ over possible states with each agent $B$. 
\begin{definition}
A {\em state partition} $\Pi$ is a collection of subsets of $2^{\mathbf F}$ that is collectively exhaustive and mutually exclusive.  For any state $s\in 2^{\mathbf{F}}$, let $\Pi(s)$ denote the element of $\Pi$ that contains $s$.
\end{definition}
If $\Pi=\{2^{\mathbf F}\}$ then we call $\Pi$ the {\em trivial partition} with respect to ${\mathbf F}$.  If $\Pi=\{\{s\} \mid s\in2^{\mathbf F}\}$, then we call $\Pi$ the {\em unit partition}.
\begin{definition}
For each $A\in\mathbf{A}$ the {\em trust function} $T_A$ is a function that maps each $B\in{\mathbf A}$ to a state partition $\Pi^B_A$.
\end{definition}
The partition $\Pi^B_A$ represents the trust that $A$ has in $B$ over different aspects of knowledge.  Informally, the partition encodes states that $A$ will trust $B$ to distinguish.  If $\Pi^B_A(s_1)\ne\Pi^B_A(s_2)$, then $A$ will trust that $B$ can distinguish between states $s_1$ and $s_2$.  Conversely, if $\Pi^B_A(s_1)=\Pi^B_A(s_2)$, then $A$ does not see $B$ as an authority capable of distinguishing between $s_1$ and $s_2$.  We clarify by returning to our motivating example.

\begin{example}
Let $\mathbf{A}=\{A,D,J\}$ and let $\mathbf{F}=\{sick, diam\}$.  Informally, the fluent $sick$ is true if $A$ has an illness and the fluent $diam$ is true if a certain piece of jewelry that $A$ is wearing contains a real diamond.  If we imagine that $D$ represents a doctor and $J$ represents a jeweler, then we can use state partitions to represent the trust that $A$ has in $D$ and $J$ with respect to different domains.  Following standard shorthand notation, we represent a state $s$ by the set of fluent symbols that are {\em true} in $s$.  In order to make the descriptions of a partition more readable, we use a $\mid$ symbol to visually separate different cells.  The following partitions are then intuitively plausible in this example:
\begin{eqnarray*}
\Pi^D_A &:=& \{sick,diam\},\{sick\} \mathbf{\mid} \{diam\},\emptyset\\
\Pi^J_A &:=& \{sick,diam\},\{diamond\} \mathbf{\mid} \{sick\},\emptyset 
\end{eqnarray*}
Hence, $A$ trusts the doctor $D$ to distinguish between states where $A$ is sick as opposed to states where $A$ is not sick.  However, $A$ does not trust $D$ to distinguish between worlds that are differentiated by the authenticity of a diamond.  The formula $sick\wedge\neg diamond$ encodes the doctor's statement that the agent is sick, and the necklace they are wearing has a fake diamond. 
\end{example}

Although the preceding example is simple, it illustrates how a partition can be used to encode the perceived expertise of agents.  In the doctor-jeweler example, we could equivalently have defined trust with respect to the set of fluents.  In other words, we could have simply said that $D$ is trusted over the fluent $sick$.  However, there are many practical cases where this is not sufficient;  we do not want to rely on the fluent vocabulary to determine what is a valid feature with respect to trust.  For example, a doctor may have specific expertise over lung infections for those working in factories, but not for lung infections for those working in a space shuttle.  By using state partitions to encode trust, we are able to capture a very flexible class of distinct areas of trust.

\subsection{Incorporating Trust in Belief Revision}
As indicated previously, we assume each agent $A$ has an AGM belief revision operator $*_A$ for incorporating new information.  In this section, we describe how the revision operator $*_A$ is combined with the trust function $T_A$ to define a new, trust-incorporating revision operator $*^B_A$.  In many cases, the operator $*^B_A$ will not be an AGM operator because it will fail to satisfy the AGM postulates.  In particular, $A$ will not necessarily believe a new formula when it is reported by an untrusted source.  This is a desirable feature.

Our approach is to define revision as a two-step process.  First, the agent considers the source and the relevant state partition to determine how much of the new information to incorporate.  Second, the agent performs standard AGM revision using the faithful assignment corresponding to the belief revision operator.
\begin{definition}
Let $\phi$ be a formula and let $T_A(B)=\Pi_A^B$.  Define:
$$\Pi_A^B[\phi] = \bigcup\{\Pi_A^B(s) \mid s\models\phi\}.$$
\end{definition}
Hence $\Pi_A^B[\phi]$ is the union of all cells that contain a model of $\phi$.  

If $A$ does not trust $B$ to distinguish between states $s$ and $t$, then any report from $B$ that provides evidence that $s$ is the actual state is also evidence that $t$ is the actual state.  When $A$ performs belief revision, it should be with respect to the distinctions that $B$ can be trusted to make.  It follows that $A$ need not believe $\phi$ after revision; instead $A$ should interpret $\phi$ to be evidence of any state $s$ that is $B$-indistinguishable from a model of $\phi$.  Formally, this means that the formula $\phi$ is construed to be evidence for each state in $\Pi_A^B[\phi]$.
\begin{definition}\label{def:trrev}
Let $A,B\in\mathbf{A}$ with $T_A(B)=\Pi^B_A$, and let $*_A$ be an AGM revision operator for $A$.  For any belief set $K$ with corresponding ordering $\prec_K$ given by the underlying faithful assignment, the trust-sensitive revision  $K*^B_A\phi$ is the set of formulas true in
$$\min_{\prec_K}(\{s \mid s\in\Pi_A^B[\phi]\}).$$
\end{definition}
So rather than taking the minimal models of $\phi$, we take all minimal states that $B$ can not be trusted to distinguish from the minimal models of $\phi$.

It is worth remarking that this notion can be formulated synactically as well.  Since ${\mathbf F}$ is finite, each state $s$ is defined by a unique, maximal conjunction over literals in ${\mathbf F}$; we simply take the conjunction of all the atomic formulas that are true in $s$ together with the negation of all the atomic formulas that are false in $s$.
\begin{definition}
For any state $s$, let $prop(s)$ denote the unique, maximal conjunction of literals true in $s$.
\end{definition}
This definition can be extended for a cell in a state partition.
\begin{definition}
Let $\Pi$ be a state partition.  For any state $s$, 
$$prop(\Pi(s)) = \bigvee \{prop(s^{\prime}) \mid  s^{\prime}\in\Pi(s)\}.$$
\end{definition}
Note that $prop(\Pi(s))$ is a well-defined formula in disjunctive normal form, due to the finiteness of ${\mathbf F}$.  Intuitively,
$prop(\Pi(s))$ is the formula that defines the partition $\Pi(s)$.  In the case of a trust partition $\Pi^B_A$, we can use this idea to define the {\em trust expansion} of a formula.
\begin{definition}\label{def:trustexpansion}
Let $A,B\in\mathbf{A}$ with the corresponding state partition $\Pi^B_A$, and let $\phi$ be a formula.  The {\em trust expansion} of $\phi$ for $A$ with respect to $B$ is the formula
$$\phi^B_A := \bigvee \{prop(\Pi^B_A(s)) \mid s\models\phi\}.$$
\end{definition}
Note that this is a finite disjunction of disjunctions, which is again a well defined formula.  We refer to $\phi^B_A$ as the trust expansion of $\phi$ because it is true in all states that are consistent with $\phi$ with respect to distinctions that $A$ trusts $B$ to be able to make.  It is an expansion because the set of models of $\phi^B_A$ is normally larger than the set of models of $\phi$.  The trust sensitive revision operator could equivalently be defined as the normal revision, following translation of $\phi$ to the corresponding trust expansion.

\begin{example}
Returning to our example, we consider a few different formulas for revision:
\begin{enumerate}
\item $\phi_1 = sick$
\item $\phi_2 = \neg diam$
\item $\phi_3 =  sick \wedge \neg diam$.
\end{enumerate}
Suppose that the agent initially believes that they are not sick, and that the diamond they have is real, so $K=\neg sick \wedge diam$.  For simplicity, we will assume that the underlying pre-order $\prec_K$ has only two levels: those states where $K$ is true are minimal, and those where $K$ is false are not.  We have the following results for revision
\begin{enumerate}
\item $K *_A^D \phi_1 = sick\wedge diam$
\item $K *_A^D \phi_2 = \neg sick\wedge diam$
\item $K *_A^D \phi_3 = sick\wedge diam$.
\end{enumerate}
The first result indicates that $A$ believes the doctor when the doctor reports that they are sick.  The second result indicates that $A$ essentially ignores a report from the doctor on the subject of jewelry.  The third result is perhaps the most interesting.  It demonstrates that our approach allows an agent to just incorporate a part of a formula.  Hence, even though $\phi_3$ is given as a single piece of information, the agent $A$ only incorporates the part of the formula over which the doctor is trusted.
\end{example}

\section{Formal Properties}
\subsection{Basic Results}
We first consider extreme cases for trust-sensitive revision operators.  Intuitively, if $T_A(B)$ is the trivial partition, then $A$ does not trust $B$ to be able to distinguish between any states.  Therefore, $A$ should not incorporate any new information obtained from $B$.  The following proposition makes this observation explicit.
\begin{proposition}\label{prop1}
If $T_A(B)$ is the trivial partition, then  $K*^B_A\phi = K$ for all $K$ and $\phi$.
\end{proposition}
The other extreme situation occurs when $T_A(B)$ is the unit partition, which consists of all singleton sets.  In this case, $A$ trusts $B$ to be able to distinguish between every possible pair of states.  It follows from this result that trust sensitive revision operators are not AGM revision operators.

\begin{proposition}\label{prop2}
If $T_A(B)$ is the unit partition, then  $*^B_A=*_A$.
\end{proposition}
Hence, if $B$ is universally trusted, then the corresponding trust sensitive revision operator is just the a priori revision operator for $A$.

\subsection{Refinements}
There is a partial ordering on partitions based on the notion of {\em refinement}.  We say that $\Pi_1$ is a refinement of $\Pi_2$ just in case, for each $S_1\in\Pi_1$, there exists $S_2\in\Pi_2$ such that $S_1\subseteq S_2$.  We also say that $\Pi_1$ is {\em finer} than $\Pi_2$.  In terms of trust-partitions, refinement has a natural interpretation in terms of ``breadth of trust.''  If the partition corresponding to $B$ is finer than that corresponding to $C$, it means that $B$ is trusted more broadly than $C$.  To be more precise, it means that $B$ is trusted to distinguish between all of the states that $C$ can distinguish, and possibly more.  If $B$ is trusted more broadly that $C$, it follows that a report from $B$ should give give $A$ more information.  This idea is formalized in the following proposition.
\begin{proposition}
For any formula $\phi$, if $\Pi^B_A$ is a refinement of $\Pi^C_A$, then  $|K*^B_A\phi| \subseteq  |K*^C_A\phi|$.
\end{proposition}
This is a desirable property; if $B$ is trusted over a greater range of states, then fewer states are possible after a report from $B$.

\subsection{Multiple Reports}
One natural question that arises is how to deal with multiple reports of information from different agents, with different trust partitions.  In our example, for instance, we might get a conflicting report from a jeweler with respect to the status of the necklace. In order to facilitate the discussion, we introduce a precise notion of a {\em report}.
\begin{definition}
A {\em report} is a pair $(B,\phi)$, where $B\in\mathbf{A}$ and $\phi$ is a formula.
\end{definition}
We can now extend the definition of trust senstive revision to reports in the obvious manner.  In fact, if the revising agent $A$ is clear from the context, we can use the short hand notation:
$$K * (\phi,B) = K *^B_A \phi.$$
The following definition extends the notion of revision to incorporate multiple reports.
\begin{definition}\label{def:multi}
Let $\{A\}\cup\mathbf{B}\subseteq \mathbf{A}$, and let $\Phi=\{(\phi_i,B_i) \mid i<n\}$ be a finite set of reports. 
Given $K$, $*$ and $\prec_K$, the trust-sensitive revision $K*_A \Phi$ is the set of formulas true in
$$\min_{\prec_K}(\{s \mid s\in\Pi_A^{Bi}[\phi_i]\}).$$
\end{definition}
So the trust sensitive revision for a finite set of reports from different agents is essentially the normal, single-shot revision by the conjunction of formulas.  The only difference is that we expand each formula with respect to the trust partition for a particular reporting agent.

\begin{example}
In the doctor and jeweler domain, we can consider how how an agent might incorporate a set of reports from $D$ and $J$.  
We start with the same initial belief set as before: $K=\neg sick \wedge diam$.  Consider the following reports:
\begin{enumerate}
\item $\Phi_1 = \{(sick, D), (\neg diam, D)\}$
\item $\Phi_2 = \{(sick,J),(\neg diam, J)\}$
\item $\Phi_3 = \{(sick, D), (\neg diam, J)\}$
\item $\Phi_4 = \{(sick,J),(\neg diam, D)\}$
\end{enumerate}
We have the following results following revision:
\begin{enumerate}
\item $K *_A \Phi_1 = sick\wedge diam$
\item $K *_A \Phi_2 = \neg sick\wedge \neg diam$
\item $K *_A \Phi_3 = sick\wedge \neg diam$
\item $K *_A \Phi_4 = \neg sick \wedge diam$.
\end{enumerate}
These results demonstrate how the agent $A$ essentially incorporates information from $D$ and $J$ in domains where they are trusted, and ignores information when they are not trusted.  Note that, in this case, $D$ and $J$ are trusted over disjoint sets of states.  As a result, it is not possible to have contradictory reports that are equally trusted.
\end{example}

The problem with Definition \ref{def:multi} is that the set of states in the minimization may be empty.  This occurs when multiple agents give conflicting reports, and we trust each agent on the domain.  In order to resolve this kind of conflict, we need a more expressive form of trust that allows some agents to be trusted more than others.  We introduce such a representation in the next section.

\section{Trust Pseudometrics}
\subsection{Measuring Trust}
In the previous section, we were concerned with a binary notion of trust that did not include any measure of the strength of trust held in a particular agent or domain.  Such an approach is appropriate in cases where we only receive new information from a single source, or from a set of sources that are equally reliable.  However, it is not sufficient if we consider cases where several different sources may provide conflicting information.  In such cases, we need to determine which information source is the most trust worthy with respect to the domain currently under consideration.

In the binary approach, we associated a partition of the state space with each agent. In order to capture different levels of trust, we would like to introduce a measure of the distance between two states from the perspective of a particular agent.  In other words, an agent $A$ would like to associate a distance function $d_B$ over states with each other agent $B$.  If $d_B(s,t)=0$, then $B$ can not be trusted to distinguish between the states $s$ and $t$.  On the other hand, if $d_B(s,t)$ is very large, then $A$ has a high level of trust in $B$'s ability to distinguish between $s$ and $t$.  The notion of distance that we introduce will be a {\em psuedometric} on the state space.  A pseudometric is a function $d$ that satisfies the following properties for all $x,y,z$ in the domain $X$:
\begin{enumerate}
\item $d(x,x)=0$
\item $d(x,y) = d(y,x)$
\item $d(x,z) \le d(x,y) + d(y,z)$
\end{enumerate} 
The difference between a {\em metric} and a pseudometric is that we do not require that $d(x,y)=0$ implies $x=y$ (the so-called law of indiscernables).  This would be undesirable in our setting, because we want to use the distance 0 to represent states that are indistinguishable rather than identical.  The first two properties are clearly desirable for a measure of our trust in another agent's ability to discern states.  The third property is the triangle inequality, and it is required to guarantee that our trust in other agents is transititive across different domains.

\begin{definition}
For each $A\in\mathbf{A}$, a {\em pseudometric trust function} ${\cal T}_A$ is a function that maps each $B\in{\mathbf A}$ to a pseudometric $d_B$ over $2^{\mathbf F}$.
\end{definition}
The pair $(2^{\mathbf F}, {\cal T}_A)$ is called a pseudometric trust space.  We would like to model the situation where a sequence of formulas $\Phi = \phi_1,\dots,\phi_n$ is received from the agents $\mathbf{B}=B_1,\dots,B_n$, respectively.  Note that the order does not matter, we think of the formulas as arriving at the same instant with no preference between them other than the preference induced by the pseudometric trust space.

We associate a sequence of state partitions with each pseudometric trust space.
\begin{proposition}
Let $(2^{\mathbf F}, {\cal T}_A)$ be a pseudometric trust space, let $B\in\mathbf{A}-A$, and let $i$ be a natural number.  For each state $s$, define the set
$\Phi_B^A(i)(s)$ as follows:
$$\Pi_B^A(i)(s) = \{t\mid d_B(s,t)\le i\}.$$
The collection of sets $\{\Pi_B^A(i)(s) \mid s\in 2^{\mathbf F}\}$ is a state partition.  
\end{proposition}
We let $\Pi_B^A(i)$ denote the state partition obtained from this proposition.  The cells of the partition $\Pi_B^A(i)$ consist of all states are separated by a distance of no more than $i$.  The following proposition is immediate.
\begin{proposition}
$\Pi_B^A(i)$ is a refinement of $\Pi_B^A(j)$, for any $i<j$.
\end{proposition}
Hence, a pseudometric trust space defines a sequence of partitions for each agent.  This sequence of partitions gets coarser as we increase the index; increasing the index corresponds to requiring a higher level of trust that an agent can distinguish between states.  Since we can use Definition \ref{def:trrev} to define a trust sensitive revision operator from a state partition, we can now define a trust sensitive revision operator for any fixed distance $i$ between states.  Informally, as $i$ increases, we require $B$ to have a greater degree of certainty in order to trust them to distinguish between states.  However, it is not clear in advance exactly which $i$ is the right threshold.  Our approach will be to find the lowest possible threshold that yields a consistent result.

Note that $\Pi_B^A(i)$ will be a trivial partition for any $i$ that is less than the minimum distance assigned by the underlying pseudometric trust function.
\begin{definition}
Let $(2^{\mathbf F}, {\cal T}_A)$ be a pseudometric trust space, and let $m$ be the least natural number such that $\Pi_B^A(m)$ is non-trival.  The 
{\em trust sensitive revision operator} for $A$ with respect to $B$  is the trust sensitive revision operator given by $\Pi_B^A(m)$.
\end{definition}
This is a simple extension of our approach based on state partitions.  In the next section, we take advantage of the added expressive power of pseudometrics.

\begin{example}
We modify the doctor example.  In order to consider different levels of trust, it is more interesting to consider a domain involving two doctors: a general practitioner $D$ and a specialist $S$.  We also assume that the vocabulary includes two fluents: $ear$ and $skin$.  Informally, $ear$ is understood to be true if the patient has an ear infection, whereas $skin$ is true if the patient has skin cancer.  The important point is that an ear infection is something that can easily be diagnosed by any doctor, whereas skin cancer is typically diagnosed by a specialist.  In order to capture these facts, we define two pseudometrics $d_D$ and $d_S$.  For simplicity, we label the possible states as follows:
\begin{eqnarray*}
s_1 &=& \{ear,skin\}\\
s_2 &=& \{ear\}\\
s_3 &=& \{skin\}\\
s_4 &=& \emptyset
\end{eqnarray*}
We define the pseudometrics as follows:
\begin{table}[hbtp]
\centering
\begin{tabular}{|c|c|c|c|c|c|c|c|} \hline
	& $s_1,s_2$ & $s_1,s_3$ &  $s_1,s_4$ &  $s_2,s_3$ &  $s_2,s_4$ & $s_3,s_4$ \\ \hline 
$d_D$ 	& 1 & 2  & 2 & 2 & 2 & 1  \\ \hline 
$d_S$ 	& 2 & 2  & 2 & 2 & 2 & 2  \\ \hline 

\end{tabular}
\end{table}
With these pseudometrics, it is easy to see that both $D$ and $S$ can distinguish all of the states.  However, $S$ is more trusted to distinguish between states related to a skin cancer diagnosis.  In our framework, we would like to ensure that this implies $S$ will be trusted in the case of conflicting reports from $D$ and $S$ with respect to skin cancer.
\end{example}

\subsection{Multiple Reports}
We view the distances in a pseudometric trust space as absolute measurements.  As such, if $d_B(s,t)>d_C(s,t)$, then we have greater trust in $B$ as opposed to $C$ as far as the ability to discern the states $s$ and $t$ is concerned.  We would like to use this intuition to resolve conflicting reports between agents.
\begin{proposition}
Let $\{A\}\cup\mathbf{B}\subseteq \mathbf{A}$, and let $\Phi=\{(\phi_i,B_i) \mid i<n\}$ be a finite set of reports.  There exists a natural number $m$
such that 
$$\bigcap_{i<n}(\Pi_A^{Bi}[\phi_i](m)) \ne \emptyset.$$
\end{proposition}
Hence, for any set of reports, we can get a non-intersecting intersection if we take a sufficiently coarse state partition.  In many cases this partition will be non-trival.  Using this proposition, we define multiple report revision as follows.
\begin{definition}\label{def:multitr}
Let $(2^{\mathbf F}, {\cal T}_A)$ be a pseudometric trust space, let $\Phi=\{(\phi_i,B_i) \mid i<n\}$ be a finite set of reports, and let $m$ be the least natural number such that  $\bigcap_{i<n}(\Pi_A^{Bi}[\phi_i](m)) \ne \emptyset.$  Given $K$, $*$ and $\prec_K$, the trust-sensitive revision $K*^{\mathbf B}_A \Phi$ is the set of formulas true in
$$\min_{\prec_K}(\{s \mid s\in\Pi_A^{Bi}[\phi_i](m)\}).$$
\end{definition}
Hence, trust-sensitive revision in this context involves finding the finest possible partition that provides a meaningful combination of the reports, and then revising with the corresponding state partition.

\section{Trust and Deceit}
To this point, we have only been concerned with modeling the trust that one agent holds in another due to perceived knowledge or expertise.  Of course, the issue of trust also arises in cases where one agent suspects that another may be dishonest.  However, the manner in which trust must be handled differs greatly in this context.  If $A$ does not trust $B$, then there is little reason for $A$ to believe any part of a message sent directly from $B$.

\section{Discussion}
\subsection{Related Work}
We are not aware of any other work on trust that explicitly deals with the interaction between trust and formal belief revision operators.  There is, however, a great deal of work on frameworks for modelling trust.  As noted previously, the focus of such work is often on building reputations.  One notable approach to this problem with an emphasis on knowledge representation is \cite{Wang07}, in which trust is built based on evidence.  This kind of approach could be used as a precursor step to build a trust metric, although one would need to account for domain expertise.

Different levels of trust are treated in \cite{Krukow07}, where a lattice structure is used to represent various levels of trust strength.  This is similar to our notion of a trust pseudometric, but it permits incomparable elements.  There are certainly situations where this is a reasonable advantage.  However, the emphasis is still on the representation of trust in {\em an agent} as opposed to trust in an agent with respect to a domain.

One notable approach that is similar to ours is the semantics of trust presented in \cite{Krukow07}, which is a domain-based approach to differential trust in an agent.  The emphasis there is on {\em trust management}, however.  That is, the authors are concerned with how agents maintain some record of trust in the other agents; they are not concerned with a differential approach to belief revision.

\subsection{Conclusion} 
In this paper, we have developed an approach to trust sensitive belief revision in which an agent is trusted only with respect to a particular domain.  This has been formally accomplished first by using state partitions to indicate which states an agent can be trusted to distinguish, and then by using distance functions to quantify the strength of trust.  In both cases, the model of trust is used as sort of a precursor to belief revision.  Each agent is able to perform belief revision based on a pre-order over states, but the actual formula for revision is parametrized and expanded based on the level of trust held in the reporting agent.

There are many directions for future work, in terms of both theory and applications.  As noted previously, one of the subtle distinctions that must be addressed is the difference between trusted {\em expertise} and trusted {\em honesty}.  The present framework does not explicitly deal with the problem of deception or {\em belief manipulation} \cite{Hunter13}; it would be useful to explore how models of trust must differ in this context.  In terms of applications, our approach could be used in any domain where agents must make decisions based on beliefs formulated from multiple reports.  This is the case, for example, in many networked communication systems.

\bibliographystyle{aaai}
\bibliography{action}

\end{document}